\documentclass[runningheads]{llncs}
\usepackage{graphicx}
\usepackage{tikz}
\usepackage{comment}
\usepackage{amsmath,amssymb} % define this before the line numbering.
\usepackage{color}

\usepackage[width=122mm,left=12mm,paperwidth=146mm,height=193mm,top=12mm,paperheight=217mm]{geometry}

\usepackage{multirow}
\usepackage{booktabs}
\usepackage{enumitem}
\usepackage[pagebackref=true,breaklinks=true,letterpaper=true,colorlinks,bookmarks=false]{hyperref} % 提供引用跳转
\usepackage[ruled,vlined,linesnumbered]{algorithm2e}
\usepackage{pifont}
\usepackage{float}
\usepackage{array}
\newcolumntype{b}{!{\vrule width 0.8pt}}
\newcolumntype{B}{!{\vrule width 1pt}}

\begin{document}
% \linenumbers
\pagestyle{headings}
\mainmatter
\def\ECCVSubNumber{4959}  % Insert your submission number here

\title{Hierarchical Dynamic Filtering Network for RGB-D Salient Object Detection} % Replace with your title

\begin{comment}
\titlerunning{ECCV-20 submission ID \ECCVSubNumber}
\authorrunning{ECCV-20 submission ID \ECCVSubNumber}
\author{Anonymous ECCV submission}
\institute{Paper ID \ECCVSubNumber}
\end{comment}
%******************

% \begin{comment}
\titlerunning{Hierarchical Dynamic Filtering Network}
\author{
 Youwei Pang\inst{1} \and
 Lihe Zhang\inst{1}\thanks{Corresponding author: zhanglihe@dlut.edu.cn} \and
 Xiaoqi Zhao\inst{1} \and
 Huchuan Lu\inst{1,2}
}
\authorrunning{Y. Pang et al.}
\institute{Dalian University of Technology, China \\
 \email{\{lartpang, zxq\}@mail.dlut.edu.cn, \{zhanglihe, lhchuan\}@dlut.edu.cn} \\
 \and Peng Cheng Laboratory
}
% \end{comment}
%******************
\maketitle

\begin{abstract}
 The main purpose of RGB-D salient object detection (SOD) is how to better integrate and utilize cross-modal fusion information. In this paper, we explore these issues from a new perspective. We integrate the features of different modalities through densely connected structures and use their mixed features to generate dynamic filters with receptive fields of different sizes. In the end, we implement a kind of more flexible and efficient multi-scale cross-modal feature processing, i.e. dynamic dilated pyramid module. In order to make the predictions have sharper edges and consistent saliency regions, we design a hybrid enhanced loss function to further optimize the results. This loss function is also validated to be effective in the single-modal RGB SOD task. In terms of six metrics, the proposed method outperforms the existing twelve methods on eight challenging benchmark datasets. A large number of experiments verify the effectiveness of the proposed module and loss function. Our code, model and results are available at \url{https://github.com/lartpang/HDFNet}.
\end{abstract}

\section{Introduction}

Salient object detection (SOD) aims to model the mechanism of human visual attention and mine the most salient objects or regions in data such as images or videos. SOD has been widely applied in many computer vision tasks, such as scene classification~\cite{sceneclassification}, video segmentation~\cite{VideoSaliencyFDP}, semantic segmentation~\cite{WSSemanticSegmentation}, foreground map evaluation~\cite{Smeasure,Emeasure} visual tracking~\cite{tracking}, person re-identification~\cite{Reid} and so on.

With the advent of the fully convolutional network~\cite{FCN}, deep learning-based SOD models \cite{DSS,PiCANet} have made great progress. Some methods~\cite{EGNet,MINet,F3Net,GateNet} have achieved very good performance on the existing benchmark datasets. However, these works are mainly based on RGB data.  They still face severe challenges when handling the cluttered or low-contrast scenes.
Recently, some works~\cite{DES,DCMC,CDCP,DF,CTMF,PCANet,MMCI} introduce the depth data as an aid to further improve the detection performance.
The depth information can more intuitively express spatial structures of the objects in a scene and provide a powerful supplement for the detection and recognition of salient objects. Using complementary modal cues, the scene can be further deeply and intelligently understood. However, limited by the way of using the depth information, RGB-D salient object detection is still great challenging.

\begin{figure*}[t]
 \begin{center}
  \includegraphics[width=\textwidth]{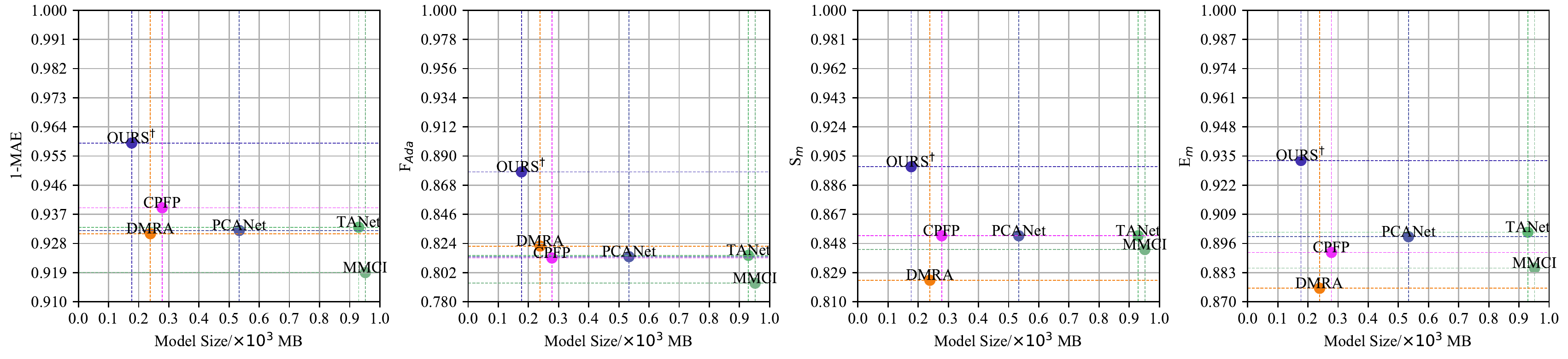}
  \caption{Comparisons in model size and accuracy.}
  \label{fig:CompParams}
 \end{center}
\end{figure*}

It is well known that RGB images contain rich appearance and detail information while depth images contain more spatial structure information. They complement each other for many vision tasks.  RGB-D SOD approaches aim to formulate cross-modal fusion in different manners. Most of them integrate depth and RGB features by element-wise addition~\cite{MMCI,DUTRGBD}, concatenation~\cite{SIP,TANet} and convolution operations~\cite{PCANet,AFNetRGBD}.
Some methods compute attention map~\cite{CPFP} or saliency map~\cite{AFNetRGBD} via a shallow or deep CNN network from pure depth images. Because of using the fixed parameters for different samples during the testing phase, the generalization capability of these models is weakened.

Moreover, for dense prediction task, the loss in each spatial position is usually different.
%Thus, the optimal gradient directions across different positions are also varying.
Thus, the actual optimization direction of gradients in different positions may be varying.
The weight-sharing convolution operation across different positions, which is used in the existing methods, causes that the training process of each parameter relies on the global gradient.
This forces the network to learn trade-off and sub-optimal parameters.
To address these problems, we propose a dynamic dilated pyramid module (DDPM), which uses RGB-depth mixed features to adaptively adjust convolution kernels for different input samples and processing locations.
These kernels can capture rich semantic cues at multiple scales with the help of the pyramid structure and the dilated convolution.
This design is capable of making more efficient convolution operations for current RGB features and promotes the network to obtain more flexible and targeted features for saliency prediction.
Early deep learning-based SOD models~\cite{CTMF}, which use fully connected layers, destroy the spatial structure of the data. This issue is alleviated to some extent by using the fully convolutional network. But the intrinsic gridding operation and the repeated down-sampling lead to the loss of numerous details in the predicted results. Although many methods frequently combine shallower features to restore feature resolution, the improvement is still limited. While some approaches~\cite{DSS,R3Net} leverage CRF post-processing to refine subtle structures, which has a large computational cost. In this work, we design a new hybrid enhanced loss function (HEL). The HEL encourages the consistency between the area around edges and the interior of objects, thereby achieving sharper boundaries and a solid saliency area.

Our main contributions are summarized as follows:

\begin{itemize}[noitemsep, nolistsep]
 \item We propose a simple yet effective hierarchical dynamic filtering network (HDFNet) for RGB-D SOD. Especially, we provide a new perspective to utilize depth information. The depth and RGB features are combined to generate region-aware dynamic filters to guide the decoding in RGB stream.
 \item We propose a hybrid enhanced loss and verify its effectiveness in both RGB and RGB-D SOD tasks. It can effectively optimize the details of predictions and enhance the consistency of salient regions without additional parameters.
 \item We compare the proposed method with twelve state-of-the-art methods on eight datasets. It achieves the best performance under six evaluation metrics. Meanwhile, we implement a forward reasoning speed of 52 FPS on an NVIDIA GTX 1080 Ti GPU. The size of our VGG16-based model is about 170 MB (Fig.~\ref{fig:CompParams}).
\end{itemize}

\section{Related Word}

\noindent\textbf{RGB-D Salient Object Detection.} The early methods are mainly based on hand-crafted features, such as contrast~\cite{DES} and shape~\cite{LS}. Limited by the representation ability of the features, they can not cope with complex scenes. Please refer to \cite{SIP} for more details about traditional methods.
In recent years, FCN-based methods have shown great potential and some of them achieve very good performance in the RGB-D SOD task~\cite{CPFP,DUTRGBD,SIP}.
Chen and Li~\cite{PCANet} progressively combine the current depth/RGB features and the preceding fused feature by a series of convolution and element-wise addition operations to build the cross fusion modules.
Recently, they concatenate depth and RGB features and feed them into an additional CNN stream to achieve multi-level cross-modal fusion~\cite{TANet}.
Wang and Gong~\cite{AFNetRGBD} respectively build a saliency prediction stream for RGB and depth inputs and then fuse their predictions and their preceding features to obtain final prediction via several convolutional layers.
Zhao \emph{et. al}~\cite{CPFP} insert a lightweight net between adjacent encoding blocks to compute a contrast map from the depth input and use it to enhance the features from the RGB stream.
Piao \emph{et. al}~\cite{DUTRGBD} combine multi-level paired complementary features from RGB and depth streams by convolution and nonlinear operations.
Fan \emph{et. al}~\cite{SIP} design a depth depurator to remove the low-quality depth input, and for high-quality one they feed the concatenated 4-channel input into a convolutional neural network to achieve cross-modal fusion.
Different from these methods, we use the RGB-depth mixed features to generate ``adaptive'' multi-scale convolution kernels to filter and enhance the decoding features from the RGB stream.

\begin{figure}[t]
 \centering
 \includegraphics[width=\textwidth]{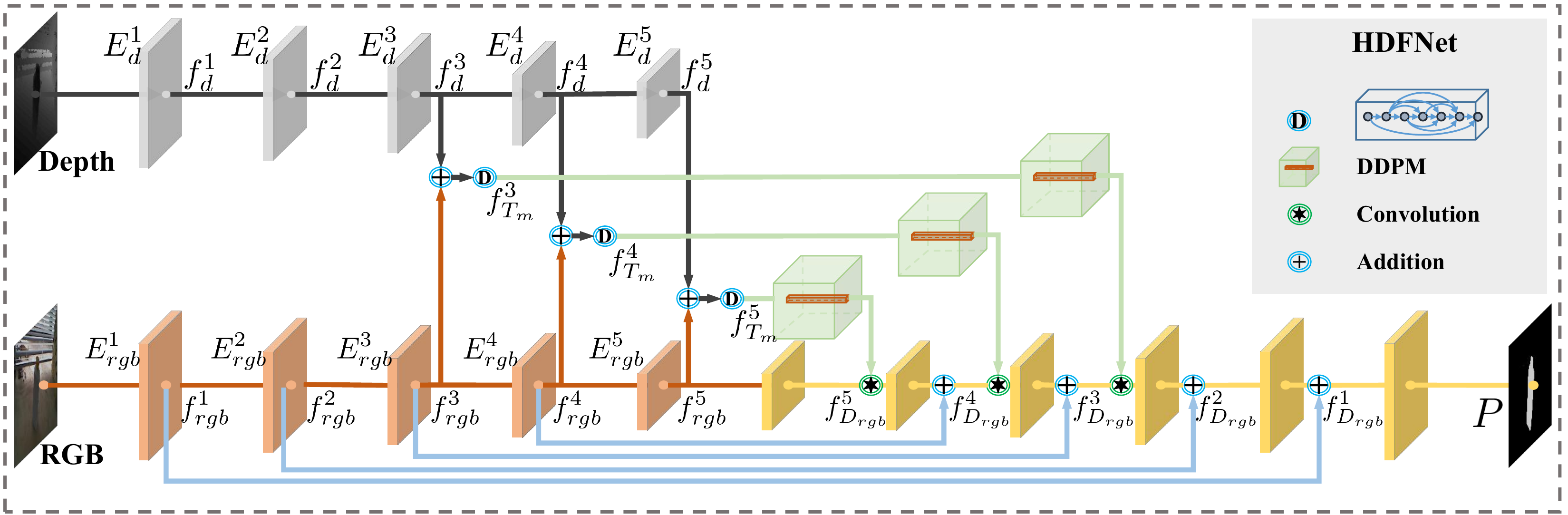}
 \caption{The overall architecture of HDFNet. The network is based on two-stream structure. The two encoders use the same network (such as VGG-16~\cite{VGG}, VGG-19~\cite{VGG}, or ResNet-50~\cite{Resnet}), and are fed RGB and depth images, respectively. The details of HDFNet are introduced in Sec.~\ref{sec:network}.}
 \label{fig:HDFNet}
\end{figure}

\noindent\textbf{Dynamic Filters.}
The works closely related to ours are \cite{DFN} and \cite{DCM}.
The conception of the dynamic filter is firstly proposed in video and stereo prediction task~\cite{DFN}. The filter is utilized to enhance the representation of its corresponding input in a self-learning manner.
While we use multi-modal information to generate multi-scale filters to dynamically strengthen the cross-modal complementarity and suppress the inter-modality incompatibility.
Besides, the kernel computation in~\cite{DFN} introduces a large number of parameters and is difficultly extended at multiple scales, which significantly increases parameters and causes optimization difficulties.
To efficiently achieve hierarchical dynamic filters, we introduce the idea of depth-wise separable convolution~\cite{MobileNet} and dilated convolution~\cite{DilatedConvolution}.
In \cite{DCM}, the filters are computed by pooling the input feature. They share kernel parameters across different positions, which is only an image-specific filter generator.
In contrast, we design position-specific and image-specific filters to provide cross-modal contextual guidance for the decoder. The parameter update of dynamic filters is determined by the gradients of local neighborhoods to achieve more targeted adjustments and guarantee the overall performance of optimization.

\section{Proposed Method}\label{sec:network}

In this section, we first introduce the overall structure of the proposed method and then detail two main components, including the dynamic dilated pyramid module (DDPM) and the hybrid enhanced loss (HEL).

\begin{figure}[t]
 \centering
 \includegraphics[width=\textwidth]{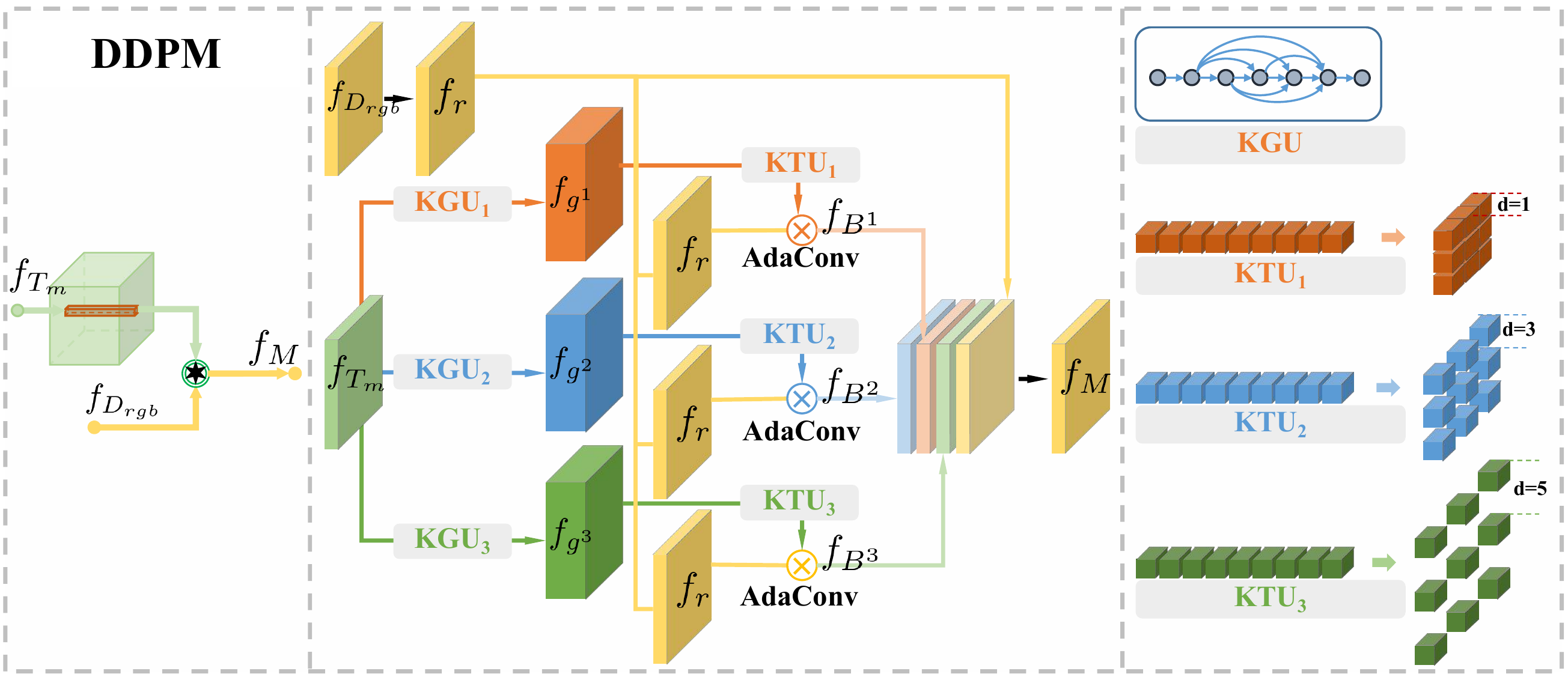}
 \caption{The structure of the dynamic dilated pyramid module. The DDPM contains two submodules: kernel generation units (KGUs) and kernel transformation units (KTUs). KGUs generate adaptive kernel tensors and KTUs transform these tensors to the regular form of convolution kernels with different dilation rates.}
 \label{fig:DDPM}
\end{figure}

\subsection{Two Stream Structure}

We build a two-stream network, which structure is shown in Fig.~\ref{fig:HDFNet}. It has two inputs: one is an RGB image and the other is a depth image, which corresponds to the RGB and depth streams, respectively.
Through convolution blocks $\{E^i_{rgb}\}^{5}_{i=1}$ and $\{E^i_{d}\}^{5}_{i=1}$ in two encoding networks, we can obtain the intermediate features with different resolutions, which are recorded as $f^{1}$, $f^{2}$, $f^{3}$, $f^{4}$, $f^{5}$ from large to small.
The third-level features still retain enough valid information. Besides, the shallower features contain more noise and also cause higher computational cost due to the larger resolution.
To balance efficiency and effectiveness, we only utilize the features $f^{3}_{d}$, $f^{4}_{d}$, $f^{5}_{d}$ from the deepest three blocks in the depth stream. These features are respectively combined with the features $f^{3}_{rgb}$, $f^{4}_{rgb}$, $f^{5}_{rgb}$ from the RGB stream.
Then, we use a dense block~\cite{DenseNet} to build the transport layer, which combines rich and various receptive fields and generates powerful mixed features $f_{T_m}$ with both spatial structures and appearance details.
These features are fed into the DDPM to produce multi-scale convolution kernels that are used to filter the features $f_{D_{rgb}}$ from the decoder. The resulted features $f_{M}$ are merged in the top-down pathway by element-wise addition. After recovering the resolution layer by layer, we obtain the final prediction $P$, which is supervised by the ground truth $G$.

\subsection{Dynamic Dilated Pyramid Module}

\begin{algorithm}[t]
 % \SetAlgoLined
 % \DontPrintSemicolon
 % $f^i_{r} = \mathcal{R}(f^i_{D_{rgb}}$
 \KwIn{$f^i_{r} = \mathcal{R}(f^i_{D_{rgb}}) \in \mathbb{R}^{N \times C' \times H' \times W'}$,  $f^i_{g^{j}} \in \mathbb{R}^{N \times (9 \times C') \times H' \times W'}$ }
 \KwOut{$f^i_{B^{j}} \in \mathbb{R}^{N \times C' \times H' \times W'}$.}
 $d \leftarrow j \times 2 - 1$\;
 pad $f^i_{r}$ with $0$ from $(H', W')$ to $(H' + 2 \times d, W' + 2 \times d)$\;
 \For{$n \leftarrow 0$ \KwTo $N - 1$}{
 \For{$c \leftarrow 0$ \KwTo $C' - 1$}{
 \For{$h \leftarrow d$ \KwTo $H' + d - 1$}{
 \For{$w \leftarrow d$ \KwTo $W' + d - 1$}{
 $(f^i_{B^{j}})_{[n,c,h,w]}$ $\leftarrow$
 $\sum^{1}_{l=-1}\sum^{1}_{m=-1} \{(f^i_{g^j})_{[n,(l+1) \times 3+(m+1),h,w]}$
 $\times (f^i_{r})_{[n,c,h+l \times d,w+m \times d]} \}$\;
 }
 }
 }
 }
 \caption{The operation process of adaptive convolution $\otimes$ related to KTU$_j$ in DDPM$^i$.}
 \label{alg:conv}
\end{algorithm}

In order to make more reasonable and effective use of the mixed features $f_{T_m}$ from the dense transport layer, we employ DDPMs to generate the adaptive kernel for decoding RGB features. The DDPMs contain two inputs: the mixed feature $f_{T_m}$ and the feature $f_{D_{rgb}}$ from the decoder.
On one hand, for specific position in feature maps $f_{D_{rgb}}$, we use kernel generation units (KGUs) to yield independent weight tensors, i.e.\ $f_{g}$, that can cover a $3 \times 3$, $7 \times 7$ or $11 \times 11$ square neighborhood. KGUs are also a kind of dense structure~\cite{DenseNet}. The module contains 4 densely connected layers and each layer is connected to all the others in a feed-forward fashion, which can further strengthen feature propagation and expression capabilities, encourage feature reuse and greatly improve parameter efficiency. Then, by recombining kernel tensors and inserting different numbers of zeros, kernel transformation units (KTUs) construct regular convolution kernels with different dilation rates. Please see ``KTU'' shown in Fig.~\ref{fig:DDPM}
and introduced in Alg.~\ref{alg:conv} for a more intuitive presentation.
On the other hand, after preliminary dimension reduction, the other input $f_{D_{rgb}}$ is re-weighted and integrated into three parallel branches to obtain the enhanced features $\{f_{B^j}\}^3_{j=1}$. Note that this is actually a channel-wise adjustment and the operation of each channel is independent.
Finally, after concating and merging $\{f_{B^j}\}^3_{j=1}$ and the reduced $f_{D_{rgb}}$, the resulted features $\{f^{i}_{M}\}^{5}_{i=3}$ become more discriminative.

The entire process can be formulated as follows:
\begin{equation}
 \begin{split}
  f^i_{M}
  & = \mathcal{DDPM}^i(f^i_{D_{rgb}}, f^i_{T_m}) \\
  & = \mathcal{F}(\mathcal{C}(\mathcal{R}(f^i_{D_{rgb}}), f^i_{B^1}, f^i_{B^2}, f^i_{B^3}) \\
  & = \mathcal{F}(\mathcal{C}(\mathcal{R}(f^i_{D_{rgb}}),
  \mathcal{KTU}^{i}_1(\mathcal{KGU}^{i}_1(f^i_{T_m})) \otimes \mathcal{R}(f^i_{D_{rgb}}), \\
  &   \quad\quad\quad \mathcal{KTU}^{i}_2(\mathcal{KGU}^{i}_2(f^i_{T_m})) \otimes \mathcal{R}(f^i_{D_{rgb}}), \\
  &   \quad\quad\quad \mathcal{KTU}^{i}_3(\mathcal{KGU}^{i}_3(f^i_{T_m})) \otimes \mathcal{R}(f^i_{D_{rgb}}))),
 \end{split}
 \label{equ:ddpm}
\end{equation}
\noindent where $f^i_{M}$ represents the feature from the DDPM$^i$ related to the $f^i_{D_{rgb}}$.
$\mathcal{DDPM}(\cdot)$, $\mathcal{KGU}(\cdot)$ and $\mathcal{KTU}(\cdot)$ denote the operation of the corresponding module.
$\mathcal{R}(\cdot)$ is a $1 \times 1$ convolution operation, which is used to reduce the number of channels from 64 to 16. $\otimes$ is an adaptive convolution operation as shown in Alg.~\ref{alg:conv}.
$\mathcal{C}(\cdot)$ is a concatenation operation and $\mathcal{F}(\cdot)$ is a $3 \times 3$ convolution to fuse the concatenated features from different branches. More details is as shown in Fig.~\ref{fig:DDPM}.

\subsection{Hybrid Enhanced Loss}

% 不论是对于RGB SOD还是RGB-D SOD任务, 良好的预测都需要显著性区域足够清晰. 这实际上包含两个方面, 一是边缘的锐利, 一是区域的一致性. 我们从损失函数入手, 分别针对边缘和前背景区域内部进行约束, 以实现高对比度的预测.
No matter for RGB or RGB-D based SOD tasks, good prediction requires the salient area to be clearly and completely highlighted. This contains two aspects: one is the sharpness of boundaries and the other is the consistency of intra-class.
We start with the loss function and design a new loss to constrain the edges and the fore-/background regions to separately achieve high-contrast predictions.

% 显著性目标检测中常用的损失函数是二值交叉熵, 主要形式如下:
The common loss function in the SOD task is binary cross entropy (BCE). It is a pixel-level loss, which independently performs error calculation and supervision at different positions. The main form is as follows:
\begin{equation}
 \begin{split}
  L_{bce} & = \frac{1}{N \times H \times W}\sum^N_n \sum^H_h\sum^W_w \left [ g \log p + (1 - g) \log (1 - p) \right ],
 \end{split}
 \label{equ:bceloss}
\end{equation}
\noindent where $P = \{ p | 0 < p < 1 \} \in \mathbb{R}^{N \times 1 \times H \times W}$ and $G = \{ g | 0 < g < 1 \} \in \mathbb{R}^{N \times 1 \times H \times W}$ respectively represent the prediction and the corresponding ground truch. $N$, $H$ and $W$ are the batchsize, height and width of the input data, respectively. It calculates the error between the ground truth $g$ and the prediction $p$ at each position, and the loss $L_{bce}$ accumulates and averages the errors of all positions.

% 为了进一步增强损失函数在边缘和区域等更高级别监督上的力度, 我们首先对边缘附近区域进行了特别的约束和优化.
In order to further enhance the strength of supervision at higher levels such as edges and regions, we specially constrain and optimize the regions near the edges. In particular, the loss is formulated as follows:
\begin{align}
 \begin{split}
  L_{e} = & \frac{\sum^{H}_{h}\sum^{W}_{w} (e * |p - g|)}{\sum^{H}_{h}\sum^{W}_{w} e}, \\
  e =     & \left \{ \begin{matrix}
   0 & \text{ if } (G - \mathcal{P}(G))_{[h,w]} = 0,   \\
   1 & \text{ if } (G - \mathcal{P}(G))_{[h,w]} \ne 0,
  \end{matrix} \right.
 \end{split}
 \label{equ:edgeloss}
\end{align}
\noindent where $L_{e}$ represents the edge enhanced loss (EEL), and $\mathcal{P}(\cdot)$ denotes the average pooling operation with a $5 \times 5$ slide window.
In Equ.~\ref{equ:edgeloss}, we can obtain the local region near the contour of the ground truth by calculating $e$. In this region, the difference $L_e$ between the prediction $p$ and the ground truth $g$ can be calculated. Through this loss, the optimization process can target the contours of salient objects.

In addition, we also design a region enhanced loss (REL)  to constrain the prediction of intra-class. By respectively calculating the prediction errors within the foreground class and the background class, fore-/background predictions can be independently optimized. Specifically, the REL $L_{r}$ is written as:
\begin{align}
 \begin{split}
  L_{r} & = \frac{\sum^{N}_{n} (L_{f} + L_{b})}{N}, \\
  L_{f} & = \frac{\sum^H_h\sum^W_w (g - g * p)}{\sum^H_h\sum^W_w g}, \\
  L_{b} & = \frac{\sum^H_h\sum^W_w (1 - g) * p}{\sum^H_h\sum^W_w (1 - g)},
 \end{split}
 \label{equ:regionloss}
\end{align}
\noindent where $L_{f}$ and $L_{b}$ denote the fore-/background losses, respectively. The losses compute the normalized prediction errors in the intra-class regions. They depict the region-level supervision.
Finally, we integrate these three losses ($L_{bce}$, $L_{e}$ and $L_{r}$) to obtain the hybrid enhanced loss (HEL), which can optimize the prediction at two different levels.  The total loss is expressed as follows:
\begin{equation}
 \begin{split}
  L = L_{bce} + L_{e} + L_{r}.
 \end{split}
 \label{equ:totalloss}
\end{equation}

\section{Experiments}

\subsection{Datasets}

To fully verify the effectiveness of the proposed method, we evaluated the results on eight benchmark datasets.
\textbf{LFSD}~\cite{LFSD} is a small dataset that contains 100 images with depth information and human-labeled ground truths and is built for saliency detection on the light filed.
\textbf{NJUD}~\cite{NLUD} contains 1,985 groups of RGB, depth, and label images, which are collected from the Internet, 3D movies, and photographs taken by a Fuji W3 stereo camera.
\textbf{NLPR}~\cite{NLPR} is also called \textbf{RGBD1000}, which contains 1,000 natural RGBD images captured by Microsoft Kinect together with the human-marked ground truth.
\textbf{RGBD135}~\cite{RGBD135} is also named \textbf{DES}, which consists of 135 images about indoor scenes collected by Microsoft Kinect.
\textbf{SIP}~\cite{SIP} includes 1,000 images with many challenging situations from various outdoor scenarios and these images emphasize salient persons in real-world scenes.
\textbf{SSD}~\cite{SSD} contains 80 images picked up from three
stereo movies.
\textbf{STEREO}~\cite{STEREO} is also called \textbf{SSB}, which contains 1,000 stereoscopic images downloaded from the Internet.
\textbf{DUTRGBD}~\cite{DUTRGBD} is a new and large dataset and contains 800 indoor and 400 outdoor scenes paired with the depth maps and ground truths.

For comprehensively and fairly evaluating different methods, we follow the setting of \cite{DUTRGBD}. On the DUTRGBD, we use 800 images for training and 400 images for testing. For the other seven datasets, we follow the data partition of \cite{PCANet,MMCI,CTMF,DUTRGBD} to use 1,485 samples from the NJUD and 700 samples from the NLPR as the training set and the remaining samples in these datasets are used for testing.

\subsection{Evaluation Metrics}

\begin{table}[t]
 \centering
 \caption{Results ($\uparrow$: $F_{max}$, $F_{ada}$~\cite{Fmeasure}, $F_{\beta}^{\omega}$~\cite{wFmeasure}, $S_{m}$~\cite{Smeasure} and $E_{m}$~\cite{Emeasure}; $\downarrow$: MAE~\cite{MAE}) of different RGB-D SOD methods across eight datasets. The best results are highlight in \textcolor{red}{\textbf{red}}. $\natural$: Traditional methods. $\dagger$: VGG-16~\cite{VGG} as backbone. $\ddagger$: VGG-19~\cite{VGG} as backbone. $\sharp$: ResNet-50~\cite{Resnet} as backbone. -: No data available.}
 \label{tab:compmodels}
 \resizebox{\textwidth}{!}{%
\begin{tabular}{@{}rl|ccccccccccc|cc|cc@{}}
   \toprule
    & \textbf{Metric}      & \textbf{DES$_{14}^{\natural}$}~\cite{DES} & \textbf{DCMC$_{16}^{\natural}$}~\cite{DCMC} & \textbf{CDCP$_{17}^{\natural}$}~\cite{CDCP} & \textbf{DF$_{17}^{\dagger}$}~\cite{DF} & \textbf{CTMF$_{18}^{\dagger}$}~\cite{CTMF} & \textbf{PCANet$_{18}^{\dagger}$}~\cite{PCANet} & \textbf{MMCI$_{19}^{\dagger}$}~\cite{MMCI} & \textbf{TANet$_{19}^{\dagger}$}~\cite{TANet} & \textbf{AFNet$_{19}^{\dagger}$}~\cite{AFNetRGBD} & \textbf{CPFP$_{19}^{\dagger}$}~\cite{CPFP} & \textbf{OURS$^{\dagger}$}       & \textbf{DMRA$_{19}^{\ddagger}$}~\cite{DUTRGBD} & \textbf{OURS$^{\ddagger}$}      & \textbf{D3Net$_{19}^{\sharp}$}~\cite{SIP} & \textbf{OURS$^{\sharp}$}        \\
   \hline
   \multirow{6}{*}{\rotatebox{90}{LFSD~\cite{LFSD}}}
    & $F_{max}$            & 0.377                                     & 0.850                                       & 0.680                                       & 0.854                                  & 0.815                                      & 0.829                                          & 0.813                                      & 0.827                                        & 0.780                                            & 0.850                                      & \textcolor{red}{\textbf{0.860}} & \textcolor{red}{\textbf{0.872}}                & 0.858                           & 0.849                                     & \textcolor{red}{\textbf{0.883}} \\
\bottomrule
  \end{tabular}%
 }
 % \vspace{-0.6cm}
\end{table}

There are six widely used metrics for evaluating RGB and RGB-D SOD models: Precision-Recall (PR) curve, F-measure~\cite{Fmeasure}, weighted F-measure~\cite{wFmeasure}, MAE~\cite{MAE}, S-measure~\cite{Smeasure} and E-measure~\cite{Emeasure}.
\noindent\textbf{PR Curve.} We use a series of fixed thresholds from 0 to 255 to binarize the gray prediction map, and then calculate several groups of precision ($Pre$) and recall ($Rec$) with ground truth by $Pre = \frac{TP}{TP+FP}$ and $Rec = \frac{TP}{TP+FN}$. Based on them, we can plot a precision-recall curve to describe the performance of the model.
\noindent\textbf{F-measure~\cite{Fmeasure}.} It is a region-based similarity metric and is formulated as the weighted harmonic mean (the weight is set to 0.3) of $Pre$ and $Rec$. In this paper, we employ the threshold changing from 0 to 255 to get $F_{max}$, and use twice the mean value of the prediction $P$ as the threshold to obtain $F_{ada}$. In addition, since F-measure reflects the performance of the binary predictions under different thresholds, we evaluate the consistency and uniformity at the regional level according to F-measure threshold curves.
\noindent\textbf{weighted F-measure ($F^{\omega}_{\beta}$)~\cite{wFmeasure}.} It is proposed to improve the existing metric F-measure. It defines a weighted precision, which is a measure of exactness, and a weighted recall, which is a measure of completeness and follows the form of F-measure.
\noindent\textbf{MAE~\cite{MAE}.} This metric estimates the approximation degree between the saliency map and ground-truth map, and it is normalized to $[0, 1]$. It focuses on pixel-level performance.
\noindent\textbf{S-measure ($S_{m}$)~\cite{Smeasure}.} It calculates the object-/region-aware structure similarities $S_{o}$ / $S_{r}$ between prediction and ground truth by the equation:
$S_{m} = \alpha \cdot S_{o} + (1 - \alpha) \cdot S_{r}, \, \alpha = 0.5$.
\noindent\textbf{E-measure ($E_{m}$)~\cite{Emeasure}.} This measure utilizes the mean-removed predictions and ground truths to compute the similarity, which characterizes both image-level statistics and local pixel matching.

\begin{figure}[tp]
 \centering
 \includegraphics[width=\textwidth]{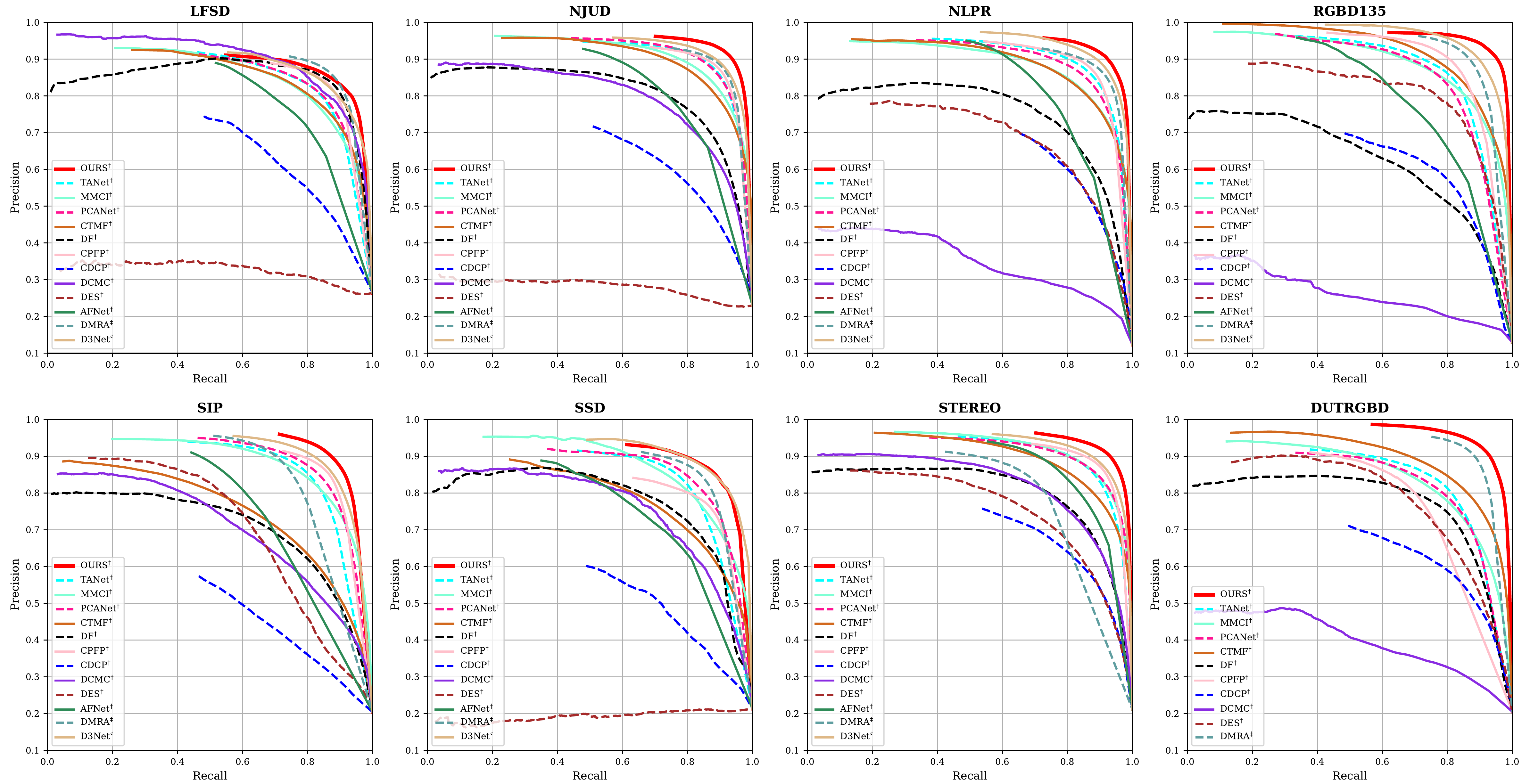}
 \caption{Precision (vertical axis) recall (horizontal axis) curves on eight RGB-D salient object detection datasets.}
 \label{fig:pr}
\end{figure}

\subsection{Implementation Details}

\noindent\textbf{Parameter setting.} Two encoders of the proposed model are based on the same model, such as VGG-16~\cite{VGG}, VGG-19~\cite{VGG}, and ResNet-50~\cite{Resnet}. In both encoders, only the convolutional layers in corresponding classification networks are retained, and the last pooling layer of VGG-16 and VGG-19 is removed at the same time. During the training phase, we use the weight parameters pre-trained on the ImageNet to initialize the encoders. Also, since the depth image is a single channel data, we change the channel number of its corresponding input layer from 3 to 1, and its parameters are initialized randomly by PyTorch. The parameters of the remaining structures are all initialized randomly.

\noindent\textbf{Training setting.} During the training stage, we apply random horizontal flipping, random rotating as data augmentation for RGB images and depth images. In addition, we employ random color jittering and normalization for RGB images. We use the momentum SGD optimizer with a weight decay of 5e-4, an initial learning rate of 5e-3, and a momentum of 0.9. Besides, we apply a ``poly'' strategy~\cite{poly} with a factor of 0.9. The input images are resized to $320 \times 320$. We train the model for 30 epochs on an NVIDIA GTX 1080 Ti GPU with a batch size of 4 to obtain the final model.

\noindent\textbf{Testing details.} During the testing stage, we resize RGB and depth images to $320 \times 320$ and normalize RGB images. Besides, the final prediction is rescaled to the original size for evaluation.

\subsection{Comparisons}

\begin{figure}[tp]
 \centering
 \includegraphics[width=\textwidth]{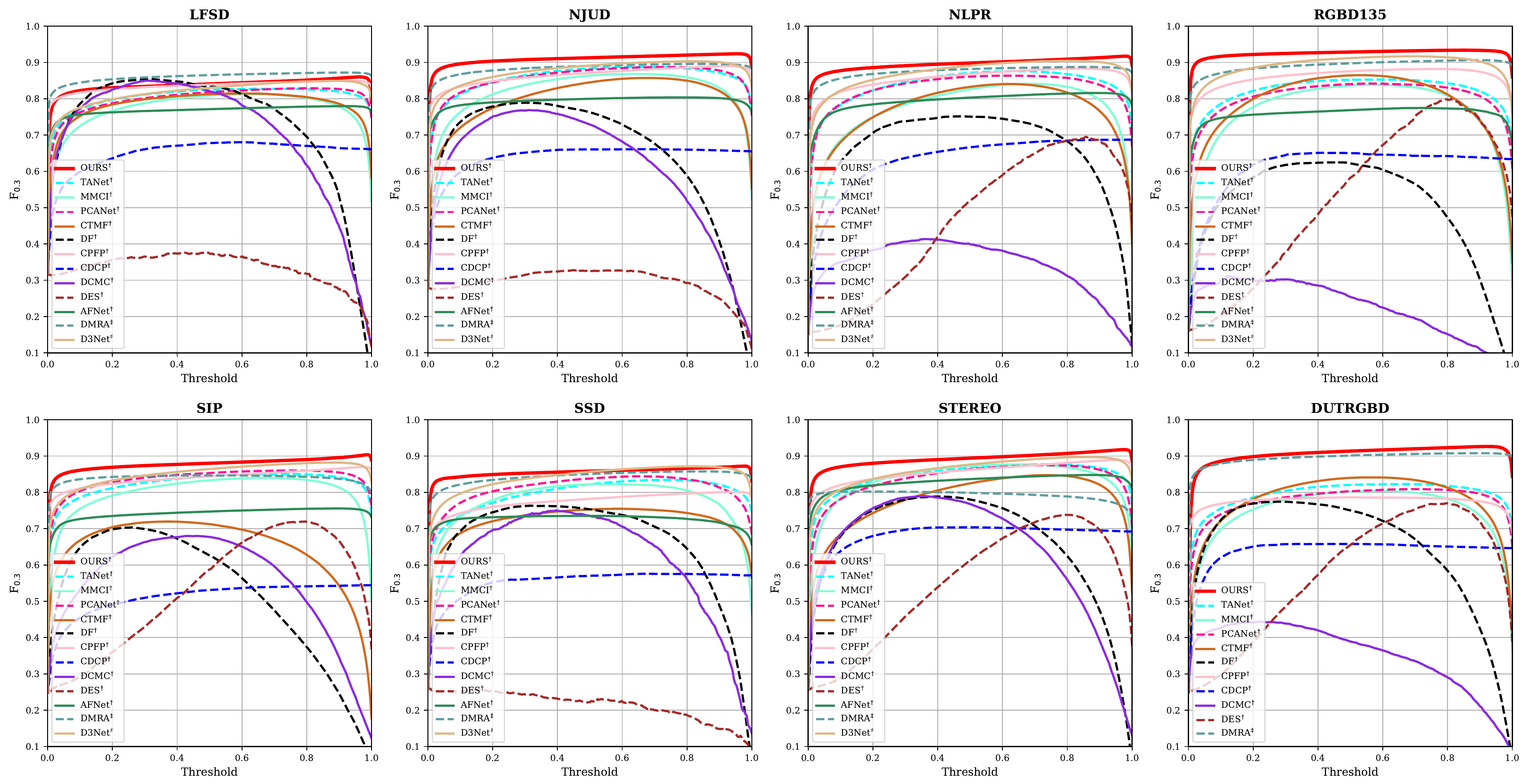}
 \caption{F-measure (vertical axis) threshold (horizontal axis) curves on eight RGB-D salient object detection datasets.}
 \label{fig:fm}
\end{figure}

In order to fully demonstrate the effectiveness of the proposed method, we compared it with the existing twelve RGB-D based SOD models, including DES~\cite{DES}, DCMC~\cite{DCMC}, CDCP~\cite{CDCP}, DF~\cite{DF}, CTMF~\cite{CTMF}, PCANet~\cite{PCANet}, MMCI~\cite{MMCI}, TANet~\cite{TANet}, AFNet~\cite{AFNetRGBD}, CPFP~\cite{CPFP}, DMRA~\cite{DUTRGBD} and D3Net~\cite{SIP}. For fair comparisons, all saliency maps of these methods are directly provided by authors or computed by their released codes. Besides, the codes and results of AFNet~\cite{AFNetRGBD} and D3Net~\cite{SIP} on the DUTRGBD~\cite{DUTRGBD} dataset are not publicly available. Therefore, their results on this dataset are not listed.

\begin{figure}[t]
 \begin{center}
  \includegraphics[width=\textwidth]{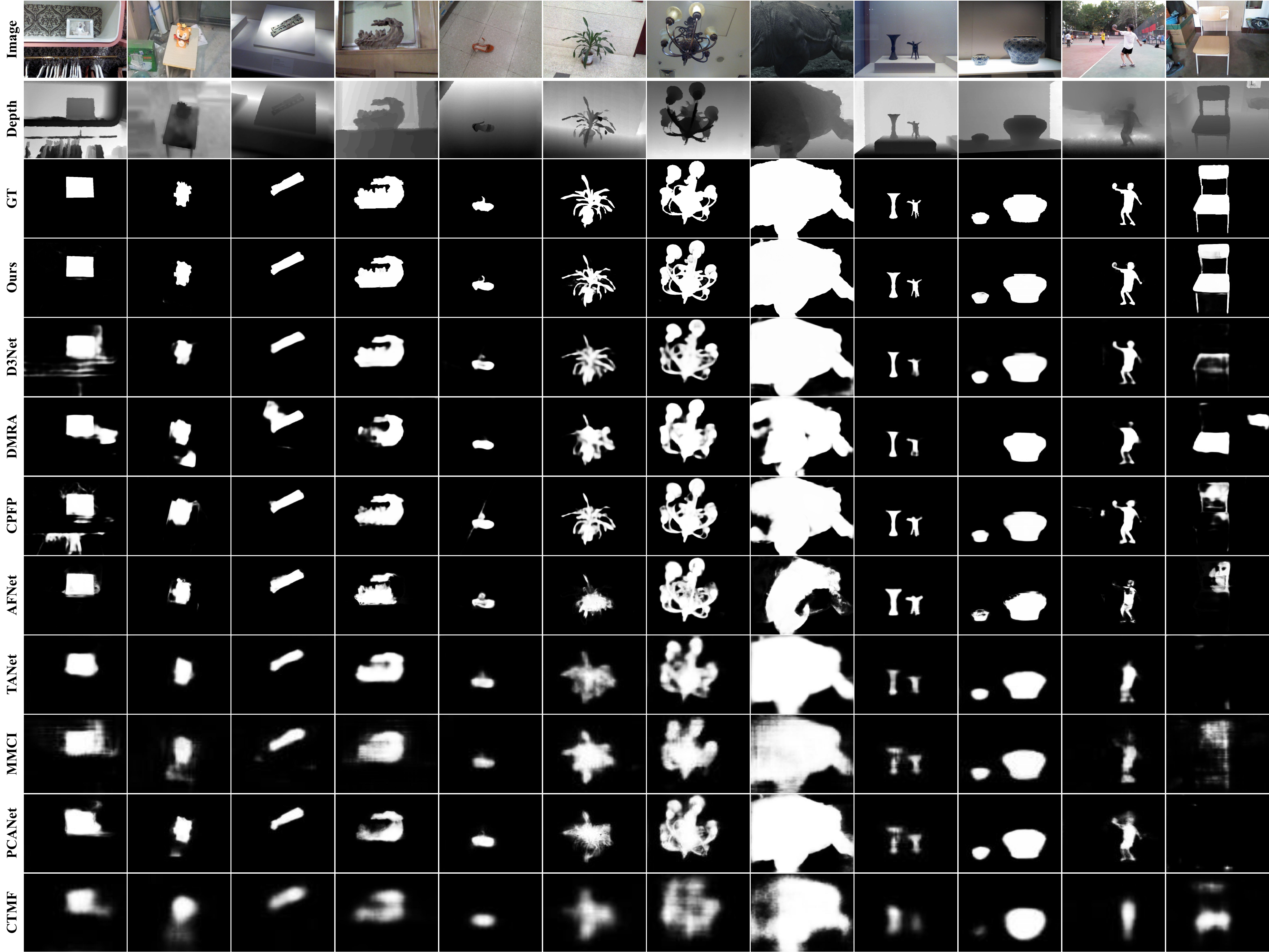}
  \caption{The visualization results of some recent methods and ours.}
  \label{fig:CompModels}
 \end{center}
\end{figure}

\noindent\textbf{Quantitative Evaluation.}
In Tab.~\ref{tab:compmodels}, we list the results of all competitors on eight datasets and six metrics. It can be seen that the proposed method performs best on most datasets and achieve significant performance improvement.
On the DUTRGBD~\cite{DUTRGBD}, our models based on VGG-16, VGG-19 and ResNet-50 have surpassed the second-best model DMRA~\cite{DUTRGBD} by 2.02\%, 2.85\% and 2.45\% on $F_{max}$, and 16.09\%, 17.88\% and 13.56\% on MAE. At the same time, on the recent dataset SIP~\cite{SIP}, they have increased by 3.83\%, 4.65\% and 5.23\% on $F_{ada}$, 5.22\%, 6.37\% and 6.84\% on $F^{\omega}_{\beta}$, and 20.94\%, 24.65\% and 24.91\% on MAE, over the D3Net~\cite{SIP}.
Because the existing RGB-D SOD datasets are relatively small, we propose a new calculation method to measure the performance of models. According to the proportion of each testing set in all testing datasets, the results on all datasets are weighted and summed to obtain an overall performance evaluation, which is listed in the row ``AveMetric'' in Tab.~\ref{tab:compmodels}.
It can be seen that our structure achieves similar and excellent results on different backbones, which shows that our structure has less dependence on the performance of the backbone.
In addition, we show a scatter plot based on the average performance of each model on all datasets and the model size in Fig.~\ref{fig:CompParams}. Our model has the smallest size while achieving the best result. We demonstrate the PR curves and the F-measure curves in Fig.~\ref{fig:pr} and Fig.~\ref{fig:fm}. Our approach (red solid line) achieves very good results on these datasets. As shown in Fig.~\ref{fig:fm}, our results are much flatter at most thresholds, which reflects that our prediction results are more uniform and consistent.

\noindent\textbf{Qualitative Evaluation.}
In Fig.~\ref{fig:CompModels}, we list some representative results. These examples include scenarios with varying complexity, as well as different types of objects, including cluttered background (Column 1 and 2), simple scene (Column 3 and 4), small objects (Column 5), complex objects (Column 6 and 7), large objects (Column 8), multiple objects (Column 9 and 10) and low contrast between foreground and background (Column 11 and 12).
It can be seen that the proposed method can consistently produce more accurate and complete saliency maps with higher contrast.

\begin{figure}[t]
 \begin{center}
  \includegraphics[width=\textwidth]{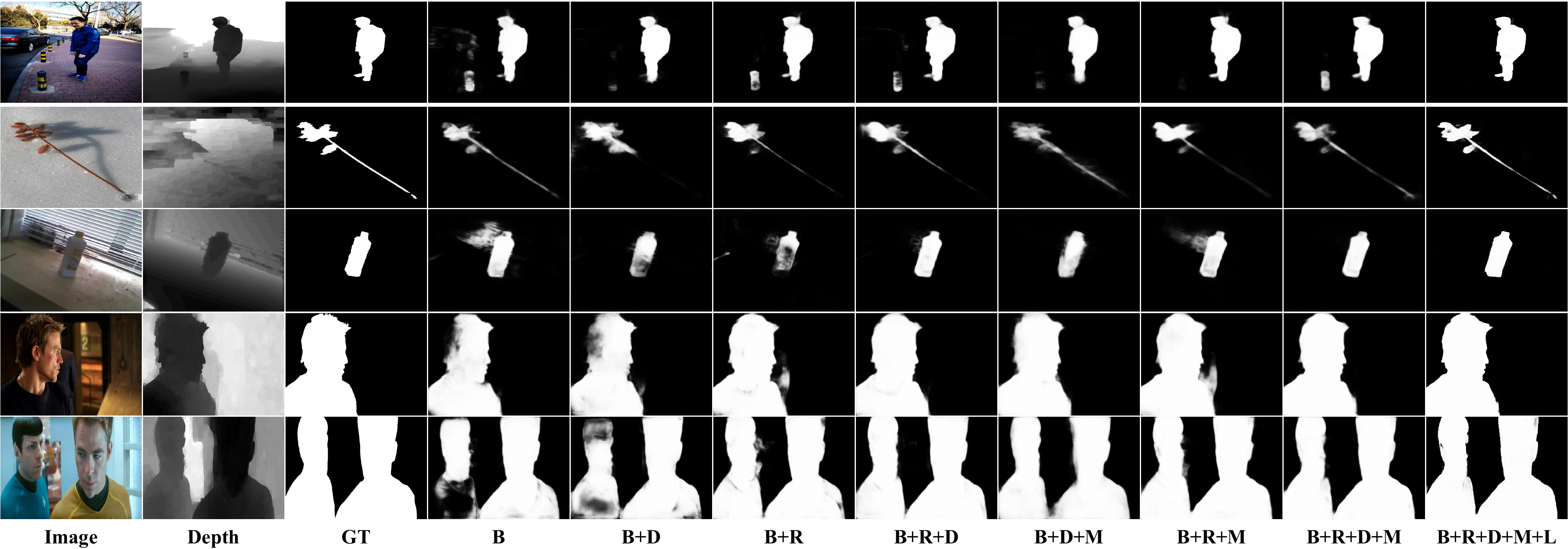}
  \caption{Visual comparisons for showing the benefits of the proposed components. GT: Ground truth; B: Baseline; D: Dense transport layer for depth features; R: Dense transport layer for RGB features; M: DDPM; L: HEL.}
  \label{fig:AblationComp}
 \end{center}
\end{figure}

\subsection{Ablation Study}

In this section, we perform ablation analysis over the main components of the HDFNet and further investigate their importance and contributions. Our baseline model, i.e.\ Model 1, uses the commonly used encoder-decoder structure, and all ablation experiments are based on the VGG-16 backbone. In the baseline model, the output features of the last three stages in the depth stream are added to the decoder after compressing the channel to 64 through an independent $1 \times 1$ convolution.
In order to evaluate the benefits of cross-modal fusion at the dense transport layer (i.e.\ Model 6), we feed single-modal features into this layer to build Model 2 (i.e.\ ``\textbf{+T$_{d}$}'') and Model 4 (i.e.\ ``\textbf{+T$_{rgb}$}''). Thus, the followed dynamic filters in the DDPM will be determined only by depth features or RGB features,  respectively.

\noindent\textbf{Dynamic Dilated Pyramid Module.}
Based on Model 2, Model 4, and Model 6, we add the dynamic dilated pyramid module to obtain Model 3, Model 5, and Model 7, respectively.
In Tab.~\ref{tab:ablation}, we show the performance improvement contributed by different structures in terms of the weighted average metrics ``AveMetric''.
It can be seen that the DDPM significantly improves performance.
Specifically, by comparing Model 3, 5 and 7 with Model 2, 4 and 6,  we achieve a relative improvement of 1.47\%, 3.11\% and 2.11\% in terms of $F^{\omega}_{\beta}$ and 5.01\%, 10.29\% and 6.77\% in terms of MAE, respectively.
We can see that even without the HEL, the average performance of Model 7 already exceeds these existing models.
More comparisons can be found in Appendix~\ref{sec:appendix}.

In addition, we compare the design of the dynamic filter in DCM~\cite{DCM} with ours.
It can be seen that the proposed DDPM (Model 7) has obvious advantages over the DCM (Model 8), and it respectively increases by 3.91\%, 5.60\%, and 18.07\% in terms of $F_{ada}$, $F^{\omega}_{\beta}$ and MAE.
In Fig.~\ref{fig:AblationComp}, we can see that the noise in depth images interferes with the final predictions. By the cross-modal guidance from the DDPMs, the interference is effectively suppressed.

\begin{table}[t]
 \centering
 \caption{Ablation experiments. \textbf{+T$_{d}$}: Using a dense transport layer for depth features. \textbf{+T$_{rgb}$}: Using a dense transport layer for RGB features. \textbf{+DDPM}: Using a DDPM after the transport layer. \textbf{+DCM}: Using the DCM~\cite{DCM} after the transport layer. \textbf{+L$_{e}$}: Using the edge loss as the auxiliary loss. \textbf{+L$_{f}$}: Using the foreground loss as the auxiliary loss. \textbf{+L$_{b}$}: Using the background loss as the auxiliary loss.}
 \label{tab:ablation}
 \resizebox{\textwidth}{!}{%
  \begin{tabular}{@{}l|c|cccccccc|cccccc@{}}
   \toprule
   \textbf{Model}                                          & \textbf{No.} & \textbf{Baseline} & \textbf{+T$_{d}$} & \textbf{+T$_{rgb}$} & \textbf{+DDPM} & \textbf{+DCM} & \textbf{+L$_{e}$} & \textbf{+L$_{f}$} & \textbf{+L$_{b}$} & $F_{max}$ & $F_{ada}$ & \textbf{$F^{\omega}_{\beta}$} & MAE   & $S_{m}$ & $E_{m}$ \\ \hline
   \multirow{12}{*}{\textbf{Ours$^{\dagger}$}}             & 1            & \ding{52}         &                   &                     &                &               &                   &                   &                   & 0.875     & 0.819     & 0.768                         & 0.067 & 0.865   & 0.898   \\ \cline{2-16}
                                                           & 2            & \ding{52}         & \ding{52}         &                     &                &               &                   &                   &                   & 0.879     & 0.820     & 0.768                         & 0.066 & 0.868   & 0.899   \\ \cline{2-16}
                                                           & 3            & \ding{52}         & \ding{52}         &                     & \ding{52}      &               &                   &                   &                   & 0.882     & 0.820     & 0.780                         & 0.063 & 0.873   & 0.900   \\ \cline{2-16}
                                                           & 4            & \ding{52}         &                   & \ding{52}           &                &               &                   &                   &                   & 0.884     & 0.839     & 0.787                         & 0.060 & 0.874   & 0.909   \\ \cline{2-16}
                                                           & 5            & \ding{52}         &                   & \ding{52}           & \ding{52}      &               &                   &                   &                   & 0.896     & 0.852     & 0.811                         & 0.054 & 0.886   & 0.916   \\ \cline{2-16}
                                                           & 6            & \ding{52}         & \ding{52}         & \ding{52}           &                &               &                   &                   &                   & 0.898     & 0.846     & 0.803                         & 0.056 & 0.884   & 0.913   \\ \cline{2-16}
                                                           & 7            & \ding{52}         & \ding{52}         & \ding{52}           & \ding{52}      &               &                   &                   &                   & 0.904     & 0.856     & 0.820                         & 0.052 & 0.893   & 0.918   \\ \cline{2-16}
                                                           & 8            & \ding{52}         & \ding{52}         & \ding{52}           &                & \ding{52}     &                   &                   &                   & 0.878     & 0.823     & 0.777                         & 0.064 & 0.871   & 0.903   \\ \cline{2-16}
                                                           & 9            & \ding{52}         & \ding{52}         & \ding{52}           & \ding{52}      &               & \ding{52}         &                   &                   & 0.909     & 0.878     & 0.849                         & 0.044 & 0.898   & 0.929   \\ \cline{2-16}
                                                           & 10           & \ding{52}         & \ding{52}         & \ding{52}           & \ding{52}      &               &                   & \ding{52}         &                   & 0.909     & 0.845     & 0.827                         & 0.050 & 0.887   & 0.916   \\ \cline{2-16}
                                                           & 11           & \ding{52}         & \ding{52}         & \ding{52}           & \ding{52}      &               &                   &                   & \ding{52}         & 0.907     & 0.874     & 0.836                         & 0.048 & 0.895   & 0.926   \\ \cline{2-16}
                                                           & 12           & \ding{52}         & \ding{52}         & \ding{52}           & \ding{52}      &               & \ding{52}         & \ding{52}         & \ding{52}         & 0.914     & 0.878     & 0.857                         & 0.041 & 0.898   & 0.933   \\ \hline
   \multirow{2}{*}{\textbf{R3Net}$_{18}$~\cite{R3Net}}     & 13           &                   &                   &                     &                &               &                   &                   &                   & 0.828     & 0.714     & 0.716                         & 0.072 & 0.831   & 0.830   \\ \cline{2-16}
                                                           & 14           &                   &                   &                     &                &               & \ding{52}         & \ding{52}         & \ding{52}         & 0.832     & 0.731     & 0.740                         & 0.069 & 0.835   & 0.844   \\ \hline
   \multirow{2}{*}{\textbf{CPD}$_{19}$~\cite{CPD}}         & 15           &                   &                   &                     &                &               &                   &                   &                   & 0.848     & 0.790     & 0.769                         & 0.052 & 0.856   & 0.889   \\ \cline{2-16}
                                                           & 16           &                   &                   &                     &                &               & \ding{52}         & \ding{52}         & \ding{52}         & 0.849     & 0.804     & 0.792                         & 0.049 & 0.857   & 0.898   \\ \hline
   \multirow{2}{*}{\textbf{PoolNet}$_{19}$~\cite{PoolNet}} & 15           &                   &                   &                     &                &               &                   &                   &                   & 0.832     & 0.755     & 0.728                         & 0.060 & 0.841   & 0.865   \\ \cline{2-16}
                                                           & 16           &                   &                   &                     &                &               & \ding{52}         & \ding{52}         & \ding{52}         & 0.861     & 0.811     & 0.799                         & 0.046 & 0.862   & 0.902   \\ \hline
   \multirow{2}{*}{\textbf{GCPANet}$_{20}$~\cite{GCPANet}} & 17           &                   &                   &                     &                &               &                   &                   &                   & 0.847     & 0.766     & 0.744                         & 0.061 & 0.854   & 0.869   \\ \cline{2-16}
                                                           & 18           &                   &                   &                     &                &               & \ding{52}         & \ding{52}         & \ding{52}         & 0.854     & 0.779     & 0.773                         & 0.055 & 0.856   & 0.880   \\ \bottomrule
  \end{tabular}%
 }
 % \vspace{-0.6cm}
\end{table}

\noindent\textbf{Hybrid Enhanced Loss.}
As shown in Tab.~\ref{tab:ablation}, the proposed hybrid enhanced loss brings huge performance improvements by comparing Model 7 with Model 12.
We evaluate each component in the HEL (Model 9, 10, and 11) and all of them contribute to the final performance.
In addition, the benefits of this loss are also clearly reflected in Fig.~\ref{fig:fm} where the curves of the proposed model are more straight, and Fig.~\ref{fig:AblationComp} where the predictions of the model ``B+R+D+M+L'' have higher contrast than ones of the model ``B+D+R+M''. Since the design goal of the HEL is to solve the general requirements of SOD tasks, we evaluate its effectiveness on several recent RGB SOD models~\cite{R3Net,CPD,PoolNet,GCPANet}. For a fair comparison, we retrain these models using the released code. Most of hyper-parameters are the same as the default values given by their corresponding code. The average performance ``AveMetric'' on five main RGB SOD datasets (DUTS~\cite{DUTS}, ECSSD~\cite{ECSSD}, HKU-IS~\cite{HKU-IS}, PASCAL-S~\cite{PASCAL-S} and DUT-OMRON~\cite{DUT-OMRON}) is shown in Tab.~\ref{tab:ablation}. More experimental details and results can be found in Appendix~\ref{sec:appendix}.

\section{Conclusions}

In this paper, we revisit the role that depth information should play in the RGB-D based SOD task.
We consider the characteristics of spatial structures contained in depth information and combine it with RGB information with rich appearance details. After that, the model generates adaptive filters with different receptive field sizes through the dynamic dilated pyramid module.
It can make full use of semantic cues from multi-modal mixed features to achieve multi-scale cross-modal guidance, thereby enhancing the representation capabilities of the decoder.
At the same time, we can obtain clearer predictions with the aid of additional region-level supervision to the regions around the edges and fore-/background regions.
Expensive experiments on eight datasets and six metrics demonstrate the effectiveness of the designed components. The proposed approach achieves state-of-the-art performance with small model size and high running speed.

\noindent\textbf{Acknowledgements.} This work was supported in part by the National Key R\&D Program of China \#2018AAA0102003, National Natural Science Foundation of China \#61876202, \#61725202, \#61751212 and \#61829102, the Dalian Science and Technology Innovation Foundation \#2019J12GX039, and the Fundamental Research Funds for the Central Universities \#DUT20ZD212.

\appendix

\section{Appendix}\label{sec:appendix}

\noindent In \textbf{Table 2} of the original paper, we show the weighted average results of each model in terms of six metrics. In Sec.~\ref{sec:app_component} and Sec~\ref{sec:app_hel} of this document, we respectively list the results of these models across different datasets.

This supplementary document is organized as follows:
\begin{itemize}[noitemsep, nolistsep]
 \item More details about the performance contributed by different components in the proposed HDFNet.
 \item More detailed comparisons of RGB SOD models with the HEL and without the HEL.
\end{itemize}

\subsection{Ablation Study}\label{sec:app_component}

Tab.~\ref{tab:app_component} shows the performance improvement contributed by different components. Note that Model 2, 4 and 6 without the DDPM directly combine the features of dense transport layer into the decoder by element-wise addition instead of using convolution operation of Model 3, 5, 7.

The experiments in Tab~\ref{tab:app_component} are divided into different groups:
\begin{enumerate}[noitemsep, nolistsep]
 \item \textbf{Model 2} vs. \textbf{Model 3}: Effectiveness of DDPM using only depth features to compute dynamic filters.
 \item \textbf{Model 4} vs. \textbf{Model 5}: Effectiveness of DDPM using only RGB features to compute dynamic filters.
 \item \textbf{Model 6} vs. \textbf{Model 7} vs. \textbf{Model 8}: Effectiveness of DDPM using two-modality features to compute dynamic filters.
 \item \textbf{Model 9} vs. \textbf{Model 10} vs. \textbf{Model 11} vs. \textbf{Model 12}: Effectiveness of three components in the HEL (L$_{e}$, L$_{f}$ and L$_{b}$) and the overall HEL.
\end{enumerate}

\subsection{Effectiveness of the HEL}\label{sec:app_hel}

Tab~\ref{tab:app_hel} shows the performance gains of the proposed loss function in some recent RGB saliency models~\cite{R3Net,CPD,PoolNet,GCPANet}. Here, ``AveMetric'' still denotes the weighted average results on all datasets and is consistent with the data in \textbf{Table 2} of the original paper.
It is worth noting that there are some differences in our experimental settings for these models. 1) R3Net~\cite{R3Net}: We use ResNeXt-101~\cite{ResNext} as the backbone as the original paper. We only change the supervision for the final prediction to the proposed HEL. 2) CPD~\cite{CPD}: ResNet-50~\cite{Resnet} is used as the backbone. For each branch, we use the proposed HEL to supervise the prediction. 3) PoolNet~\cite{PoolNet}: The backbone network is ResNet-50. We do not use the strategy of joint training with the edge and we apply the HEL on the final prediction. 4) GCPANet~\cite{GCPANet}: We also use ResNet-50, and the HEL to supervise the final result with the same resolution as the input.

\begin{table}[H]
 \centering
 \caption{Ablation experiments of different components. ``Model $\star$'' corresponds to the model ``No.~$\star$'' in \textbf{Table 2} of the original paper. In each set of comparative experiments, we emphasize the best test results in \textcolor{red}{\textbf{red}}.}
 \label{tab:app_component}
 \resizebox{\textwidth}{!}{%
\begin{tabular}{@{}lBlBcbcc|cc|cccbcccc@{}}
   \toprule
                                                           & \multirow{3}{*}{\textbf{Metric}} & \multirow{2}{*}{Baseline}
                                                           & \multicolumn{7}{cb}{DDPM}
                                                           & \multicolumn{4}{c}{HEL}                                                                                                                                                                                                                                                                                                                                                                                                  \\ \cline{4-14}
                                                           &
                                                           &
                                                           & \multicolumn{2}{c|}{Depth Input}
                                                           & \multicolumn{2}{c|}{RGB Input}
                                                           & \multicolumn{3}{cb}{RGB-D Input}
                                                           & L$_{e}$
                                                           & L$_{f}$
                                                           & L$_{b}$
                                                           & ALL                                                                                                                                                                                                                                                                                                                                                                                                                      \\
\bottomrule
  \end{tabular}%
 }
\end{table}

\begin{table}[H]
 \centering
 \caption{Comparisons of RGB SOD models with the HEL (w) and without the HEL (w/o). The best result of each group is highlight in \textcolor{red}{\textbf{red}}.}
 \label{tab:app_hel}
 \resizebox{\textwidth}{!}{%
  \begin{tabular}{l|l|cc|cc|cc|cc}
   \toprule
                                               & \multirow{2}{*}{\textbf{Metric}}
                                               & \multicolumn{2}{c|}{\textbf{R3Net$_{18}$}~\cite{R3Net}}
                                               & \multicolumn{2}{c|}{\textbf{CPD$_{19}$}~\cite{CPD}}
                                               & \multicolumn{2}{c|}{\textbf{PoolNet$_{19}$}~\cite{PoolNet}}
                                               & \multicolumn{2}{c}{\textbf{GCPANet$_{20}$}~\cite{GCPANet}}                                                                                                                                                                                                                                                        \\ \cline{3-10}
                                               &                                                             & w/o                             & w                               & w/o                             & w                               & w/o   & w                               & w/o                             & w                               \\ \hline
   \multirow{6}{*}{DUTS~\cite{DUTS}}           & $F_{max}$                                                   & 0.823                           & \textcolor{red}{\textbf{0.827}} & 0.858                           & \textcolor{red}{\textbf{0.859}} & 0.844 & \textcolor{red}{\textbf{0.879}} & 0.856                           & \textcolor{red}{\textbf{0.865}} \\
                                               & $F_{ada}$                                                   & 0.688                           & \textcolor{red}{\textbf{0.710}} & 0.786                           & \textcolor{red}{\textbf{0.807}} & 0.748 & \textcolor{red}{\textbf{0.818}} & 0.759                           & \textcolor{red}{\textbf{0.777}} \\
\bottomrule
  \end{tabular}%
 }
\end{table}

\bibliographystyle{splncs04}
\bibliography{egbib}

\begin{thebibliography}{10}
\providecommand{\url}[1]{\texttt{#1}}
\providecommand{\urlprefix}{URL }
\providecommand{\doi}[1]{https://doi.org/#1}

\bibitem{Fmeasure}
Achanta, R., Hemami, S., Estrada, F., S{\"u}sstrunk, S.: Frequency-tuned
  salient region detection. In: Proceedings of IEEE Conference on Computer
  Vision and Pattern Recognition. pp. 1597--1604. No.~CONF (2009)

\bibitem{PCANet}
Chen, H., Li, Y.: Progressively complementarity-aware fusion network for rgb-d
  salient object detection. In: Proceedings of IEEE Conference on Computer
  Vision and Pattern Recognition. pp. 3051--3060 (2018)

\bibitem{TANet}
Chen, H., Li, Y.: Three-stream attention-aware network for rgb-d salient object
  detection. IEEE Transactions on Image Processing  \textbf{28}(6),  2825--2835
  (2019)

\bibitem{MMCI}
Chen, H., Li, Y., Su, D.: Multi-modal fusion network with multi-scale
  multi-path and cross-modal interactions for rgb-d salient object detection.
  Pattern Recognition  \textbf{86},  376--385 (2019)

\bibitem{GCPANet}
Chen, Z., Xu, Q., Cong, R., Huang, Q.: Global context-aware progressive
  aggregation network for salient object detection. In: AAAI Conference on
  Artificial Intelligence (2020)

\bibitem{DES}
Cheng, Y., Fu, H., Wei, X., Xiao, J., Cao, X.: Depth enhanced saliency
  detection method. In: Proceedings of the International Conference on Internet
  Multimedia Computing and Service. pp. 23--27 (2014)

\bibitem{RGBD135}
Cheng, Y., Fu, H., Wei, X., Xiao, J., Cao, X.: Depth enhanced saliency
  detection method. In: Proceedings of the International Conference on Internet
  Multimedia Computing and Service. pp. 23--27 (2014)

\bibitem{LS}
Ciptadi, A., Hermans, T., Rehg, J.M.: An in depth view of saliency (2013)

\bibitem{DCMC}
Cong, R., Lei, J., Zhang, C., Huang, Q., Cao, X., Hou, C.: Saliency detection
  for stereoscopic images based on depth confidence analysis and multiple cues
  fusion. IEEE Signal Processing Letters  \textbf{23}(6),  819--823 (2016)

\bibitem{R3Net}
Deng, Z., Hu, X., Zhu, L., Xu, X., Qin, J., Han, G., Heng, P.A.: R3net:
  Recurrent residual refinement network for saliency detection. In:
  International Joint Conference on Artificial Intelligence. pp. 684--690
  (2018)

\bibitem{Smeasure}
Fan, D.P., Cheng, M.M., Liu, Y., Li, T., Borji, A.: Structure-measure: A new
  way to evaluate foreground maps. In: Proceedings of the IEEE International
  Conference on Computer Vision. pp. 4548--4557 (2017)

\bibitem{Emeasure}
{Fan}, D.P., {Gong}, C., {Cao}, Y., {Ren}, B., {Cheng}, M.M., {Borji}, A.:
  Enhanced-alignment measure for binary foreground map evaluation. In:
  International Joint Conference on Artificial Intelligence. pp. 698--704
  (2018)

\bibitem{SIP}
Fan, D.P., Lin, Z., Zhao, J.X., Liu, Y., Zhang, Z., Hou, Q., Zhu, M., Cheng,
  M.M.: Rethinking rgb-d salient object detection: Models, datasets, and
  large-scale benchmarks. arXiv preprint arXiv:1907.06781  (2019)

\bibitem{VideoSaliencyFDP}
Fan, D.P., Wang, W., Cheng, M.M., Shen, J.: Shifting more attention to video
  salient object detection. In: Proceedings of IEEE Conference on Computer
  Vision and Pattern Recognition. pp. 8554--8564 (2019)

\bibitem{CTMF}
Han, J., Chen, H., Liu, N., Yan, C., Li, X.: Cnns-based rgb-d saliency
  detection via cross-view transfer and multiview fusion. IEEE Transactions on
  Cybernetics  \textbf{48}(11),  3171--3183 (2017)

\bibitem{DCM}
He, J., Deng, Z., Qiao, Y.: Dynamic multi-scale filters for semantic
  segmentation. In: Proceedings of the IEEE International Conference on
  Computer Vision. pp. 3562--3572 (2019)

\bibitem{Resnet}
He, K., Zhang, X., Ren, S., Sun, J.: Deep residual learning for image
  recognition. In: Proceedings of IEEE Conference on Computer Vision and
  Pattern Recognition. pp. 770--778 (2016)

\bibitem{DSS}
Hou, Q., Cheng, M.M., Hu, X., Borji, A., Tu, Z., Torr, P.H.: Deeply supervised
  salient object detection with short connections. In: Proceedings of IEEE
  Conference on Computer Vision and Pattern Recognition. pp. 3203--3212 (2017)

\bibitem{MobileNet}
Howard, A.G., Zhu, M., Chen, B., Kalenichenko, D., Wang, W., Weyand, T.,
  Andreetto, M., Adam, H.: Mobilenets: Efficient convolutional neural networks
  for mobile vision applications. arXiv preprint arXiv:1704.04861  (2017)

\bibitem{DenseNet}
Huang, G., Liu, Z., Van Der~Maaten, L., Weinberger, K.Q.: Densely connected
  convolutional networks. In: Proceedings of IEEE Conference on Computer Vision
  and Pattern Recognition. pp. 4700--4708 (2017)

\bibitem{DFN}
Jia, X., De~Brabandere, B., Tuytelaars, T., Gool, L.V.: Dynamic filter
  networks. In: Conference and Workshop on Neural Information Processing
  Systems. pp. 667--675 (2016)

\bibitem{NLUD}
Ju, R., Liu, Y., Ren, T., Ge, L., Wu, G.: Depth-aware salient object detection
  using anisotropic center-surround difference. Signal Processing: Image
  Communication  \textbf{38},  115--126 (2015)

\bibitem{F3Net}
Jun~Wei, Shuhui~Wang, Q.H.: F3net: Fusion, feedback and focus for salient
  object detection. In: AAAI Conference on Artificial Intelligence (2020)

\bibitem{HKU-IS}
Li, G., Yu, Y.: Visual saliency based on multiscale deep features. In:
  Proceedings of IEEE Conference on Computer Vision and Pattern Recognition.
  pp. 5455--5463 (2015)

\bibitem{LFSD}
Li, N., Ye, J., Ji, Y., Ling, H., Yu, J.: Saliency detection on light field.
  In: Proceedings of IEEE Conference on Computer Vision and Pattern
  Recognition. pp. 2806--2813 (2014)

\bibitem{PASCAL-S}
Li, Y., Hou, X., Koch, C., Rehg, J.M., Yuille, A.L.: The secrets of salient
  object segmentation. In: Proceedings of IEEE Conference on Computer Vision
  and Pattern Recognition. pp. 280--287 (2014)

\bibitem{PoolNet}
Liu, J.J., Hou, Q., Cheng, M.M., Feng, J., Jiang, J.: A simple pooling-based
  design for real-time salient object detection. In: Proceedings of IEEE
  Conference on Computer Vision and Pattern Recognition (2019)

\bibitem{PiCANet}
Liu, N., Han, J., Yang, M.H.: Picanet: Learning pixel-wise contextual attention
  for saliency detection. In: Proceedings of IEEE Conference on Computer Vision
  and Pattern Recognition. pp. 3089--3098 (2018)

\bibitem{poly}
Liu, W., Rabinovich, A., Berg, A.C.: Parsenet: Looking wider to see better.
  arXiv preprint arXiv:1506.04579  (2015)

\bibitem{FCN}
Long, J., Shelhamer, E., Darrell, T.: Fully convolutional networks for semantic
  segmentation. In: Proceedings of IEEE Conference on Computer Vision and
  Pattern Recognition. pp. 3431--3440 (2015)

\bibitem{tracking}
Mahadevan, V., Vasconcelos, N.: Saliency-based discriminant tracking. In:
  Proceedings of IEEE Conference on Computer Vision and Pattern Recognition
  (2009)

\bibitem{wFmeasure}
Margolin, R., Zelnik-Manor, L., Tal, A.: How to evaluate foreground maps? In:
  Proceedings of IEEE Conference on Computer Vision and Pattern Recognition.
  pp. 248--255 (2014)

\bibitem{STEREO}
Niu, Y., Geng, Y., Li, X., Liu, F.: Leveraging stereopsis for saliency
  analysis. In: Proceedings of IEEE Conference on Computer Vision and Pattern
  Recognition. pp. 454--461 (2012)

\bibitem{MINet}
Pang, Y., Zhao, X., Zhang, L., Lu, H.: Multi-scale interactive network for
  salient object detection. In: Proceedings of IEEE Conference on Computer
  Vision and Pattern Recognition (2020)

\bibitem{NLPR}
Peng, H., Li, B., Xiong, W., Hu, W., Ji, R.: Rgbd salient object detection: a
  benchmark and algorithms. In: Proceedings of European Conference on Computer
  Vision. pp. 92--109 (2014)

\bibitem{MAE}
Perazzi, F., Kr{\"a}henb{\"u}hl, P., Pritch, Y., Hornung, A.: Saliency filters:
  Contrast based filtering for salient region detection. In: Proceedings of
  IEEE Conference on Computer Vision and Pattern Recognition. pp. 733--740
  (2012)

\bibitem{DUTRGBD}
Piao, Y., Ji, W., Li, J., Zhang, M., Lu, H.: Depth-induced multi-scale
  recurrent attention network for saliency detection. In: Proceedings of the
  IEEE International Conference on Computer Vision. pp. 7254--7263 (2019)

\bibitem{DF}
Qu, L., He, S., Zhang, J., Tian, J., Tang, Y., Yang, Q.: Rgbd salient object
  detection via deep fusion. IEEE Transactions on Image Processing
  \textbf{26}(5),  2274--2285 (2017)

\bibitem{sceneclassification}
Ren, Z., Gao, S., Chia, L.T., Tsang, I.W.H.: Region-based saliency detection
  and its application in object recognition. IEEE Transactions on Circuits and
  Systems for Video Technology  \textbf{24}(5),  769--779 (2013)

\bibitem{Reid}
Rui, Z., Ouyang, W., Wang, X.: Unsupervised salience learning for person
  re-identification. In: Proceedings of IEEE Conference on Computer Vision and
  Pattern Recognition (2013)

\bibitem{VGG}
Simonyan, K., Zisserman, A.: Very deep convolutional networks for large-scale
  image recognition. arXiv preprint arXiv:1409.1556  (2014)

\bibitem{DUTS}
Wang, L., Lu, H., Wang, Y., Feng, M., Wang, D., Yin, B., Ruan, X.: Learning to
  detect salient objects with image-level supervision. In: Proceedings of IEEE
  Conference on Computer Vision and Pattern Recognition. pp. 136--145 (2017)

\bibitem{AFNetRGBD}
Wang, N., Gong, X.: Adaptive fusion for rgb-d salient object detection. IEEE
  Access  \textbf{7},  55277--55284 (2019)

\bibitem{WSSemanticSegmentation}
Wei, Y., Liang, X., Chen, Y., Shen, X., Cheng, M.M., Feng, J., Zhao, Y., Yan,
  S.: Stc: A simple to complex framework for weakly-supervised semantic
  segmentation. IEEE Transactions on Pattern Analysis and Machine Intelligence
  \textbf{39}(11),  2314--2320 (2016)

\bibitem{CPD}
Wu, Z., Su, L., Huang, Q.: Cascaded partial decoder for fast and accurate
  salient object detection. In: Proceedings of IEEE Conference on Computer
  Vision and Pattern Recognition. pp. 3907--3916 (2019)

\bibitem{ResNext}
Xie, S., Girshick, R., Doll{\'a}r, P., Tu, Z., He, K.: Aggregated residual
  transformations for deep neural networks. In: Proceedings of IEEE Conference
  on Computer Vision and Pattern Recognition. pp. 1492--1500 (2017)

\bibitem{ECSSD}
Yan, Q., Xu, L., Shi, J., Jia, J.: Hierarchical saliency detection. In:
  Proceedings of IEEE Conference on Computer Vision and Pattern Recognition.
  pp. 1155--1162 (2013)

\bibitem{DUT-OMRON}
Yang, C., Zhang, L., Lu, H., Ruan, X., Yang, M.H.: Saliency detection via
  graph-based manifold ranking. In: Proceedings of IEEE Conference on Computer
  Vision and Pattern Recognition. pp. 3166--3173 (2013)

\bibitem{DilatedConvolution}
Yu, F., Koltun, V.: Multi-scale context aggregation by dilated convolutions.
  In: Bengio, Y., LeCun, Y. (eds.) International Conference on Learning
  Representations (2016), \url{http://arxiv.org/abs/1511.07122}

\bibitem{CPFP}
Zhao, J.X., Cao, Y., Fan, D.P., Cheng, M.M., Li, X.Y., Zhang, L.: Contrast
  prior and fluid pyramid integration for rgbd salient object detection. In:
  Proceedings of IEEE Conference on Computer Vision and Pattern Recognition.
  pp. 3927--3936 (2019)

\bibitem{EGNet}
Zhao, J.X., Liu, J.J., Fan, D.P., Cao, Y., Yang, J., Cheng, M.M.: Egnet:edge
  guidance network for salient object detection. In: Proceedings of the IEEE
  International Conference on Computer Vision (Oct 2019)

\bibitem{GateNet}
Zhao, X., Pang, Y., Zhang, L., Lu, H., Zhang, L.: Suppress and balance: A
  simple gated network for salient object detection. In: Proceedings of
  European Conference on Computer Vision (2020)

\bibitem{SSD}
Zhu, C., Li, G.: A three-pathway psychobiological framework of salient object
  detection using stereoscopic technology. In: International Conference on
  Computer Vision Workshops. pp. 3008--3014 (2017)

\bibitem{CDCP}
Zhu, C., Li, G., Wang, W., Wang, R.: An innovative salient object detection
  using center-dark channel prior. In: International Conference on Computer
  Vision Workshops. pp. 1509--1515 (2017)

\end{thebibliography}
\end{document}